\documentclass[lettersize,journal]{IEEEtran}
\usepackage{amsmath,amsfonts}
\usepackage{algorithmic}
\usepackage{array}
\usepackage[caption=false,font=normalsize,labelfont=sf,textfont=sf]{subfig}
\usepackage{textcomp}
\usepackage{stfloats}
\usepackage{url}
\usepackage{verbatim}
\usepackage{graphicx}

\usepackage{times}
\usepackage{epsfig}
\usepackage{amssymb}
\usepackage{amsthm}
\usepackage{booktabs}
\usepackage{multirow}
\usepackage{xcolor}
\usepackage{colortbl,color,framed}
\newcommand{\Sref}[1]{Section~\ref{#1}}

\newcommand{\tzt}[1]{{\color{black}{#1}}}
\newcommand{\yw}[1]{{\color{black}{#1}}}

\def\BibTeX{{\rm B\kern-.05em{\sc i\kern-.025em b}\kern-.08em
    T\kern-.1667em\lower.7ex\hbox{E}\kern-.125emX}}
\usepackage{balance}

\begin{document}
\title{Exploring the Application of Large-scale Pre-trained Models on Adverse Weather Removal}
\author{Zhentao Tan, Yue Wu$^\dagger$, Qiankun Liu, Qi Chu, Le Lu, Jieping Ye, Nenghai Yu
\IEEEcompsocitemizethanks{Zhentao Tan is with DAMO Academy, Alibaba Group and CAS Key Laboratory of Electromagnetic Space Information, University of Science and Technology of China. (Email: tanzhentao.tzt@alibaba-inc.com; tzt@mail.ustc.edu.cn). \\
Yue Wu, Le Lu and Jieping Ye are with DAMO Academy, Alibaba Group. (Email: (matthew.wy, le.lu, yejieping.ye)@alibaba-inc.com). \\
Qiankun Liu, Qi Chu and Nenghai Yu are with CAS Key Laboratory of Electromagnetic Space Information, University of Science and Technology of China. (Email: liuqk3@mail.ustc.edu.cn; (qchu, ynh)@ustc.edu.cn). \\
$^\dagger$ Yue Wu is the corresponding author.}
}


\maketitle

\begin{abstract}
Image restoration under adverse weather conditions (e.g., rain, snow and haze) is a fundamental computer vision problem and has important indications for various downstream applications. Different from early methods that are specially designed for specific type of weather, most recent works tend to remove various adverse weather effects simultaneously through either spatial feature representation learning or semantic information embedding. Inspired by the various successful applications of large-scale pre-trained models (e.g, CLIP), in this paper, \tzt{we explore the potential benefits of them for this task through both spatial feature representation learning and semantic information embedding aspects: }
1) for spatial feature representation learning, we design a Spatially-Adaptive Residual (\textbf{SAR}) Encoder to extract degraded areas adaptively. To facilitate its training, we propose a Soft Residual Distillation (\textbf{CLIP-SRD}) strategy to transfer the spatial knowledge from CLIP between clean and adverse weather images;
2) for semantic information embedding, we propose a CLIP Weather Prior (\textbf{CWP}) embedding module to make the network handle different weather conditions adaptively. This module integrates the sample specific weather prior extracted by CLIP image encoder together with the distribution specific information learned by a set of parameters, and embeds them through a cross attention mechanism. 
Extensive experiments demonstrate that our proposed method can achieve state-of-the-art performance under different and challenging adverse weather conditions.
Code will be made available.

\end{abstract}

\begin{IEEEkeywords}
Adverse Weather Removal, Image Restoration, Multi-modal Pre-trained Model, 
\end{IEEEkeywords}

\section{Introduction}
\label{sec:intro}
Images captured in daily life are commonly affected by bad weather such as rain, snow and haze, which may degrade the visual clarity of images and seriously deteriorate the performance of high-level visual applications (e.g. objection detection~\cite{pfeuffer2018optimal,liu2022image}, semantic segmentation~\cite{valada2017adapnet,pfeuffer2019robust}). Hence, removing adverse weather effects from images is essential and has been widely researched as a specific type of image restoration task~\cite{fu2017removing,qian2018attentive,wu2021contrastive,li2020deep,liu2018desnownet,ren2017video}. 

\begin{figure}[t]
    \centering
    \includegraphics[width=0.85\linewidth]{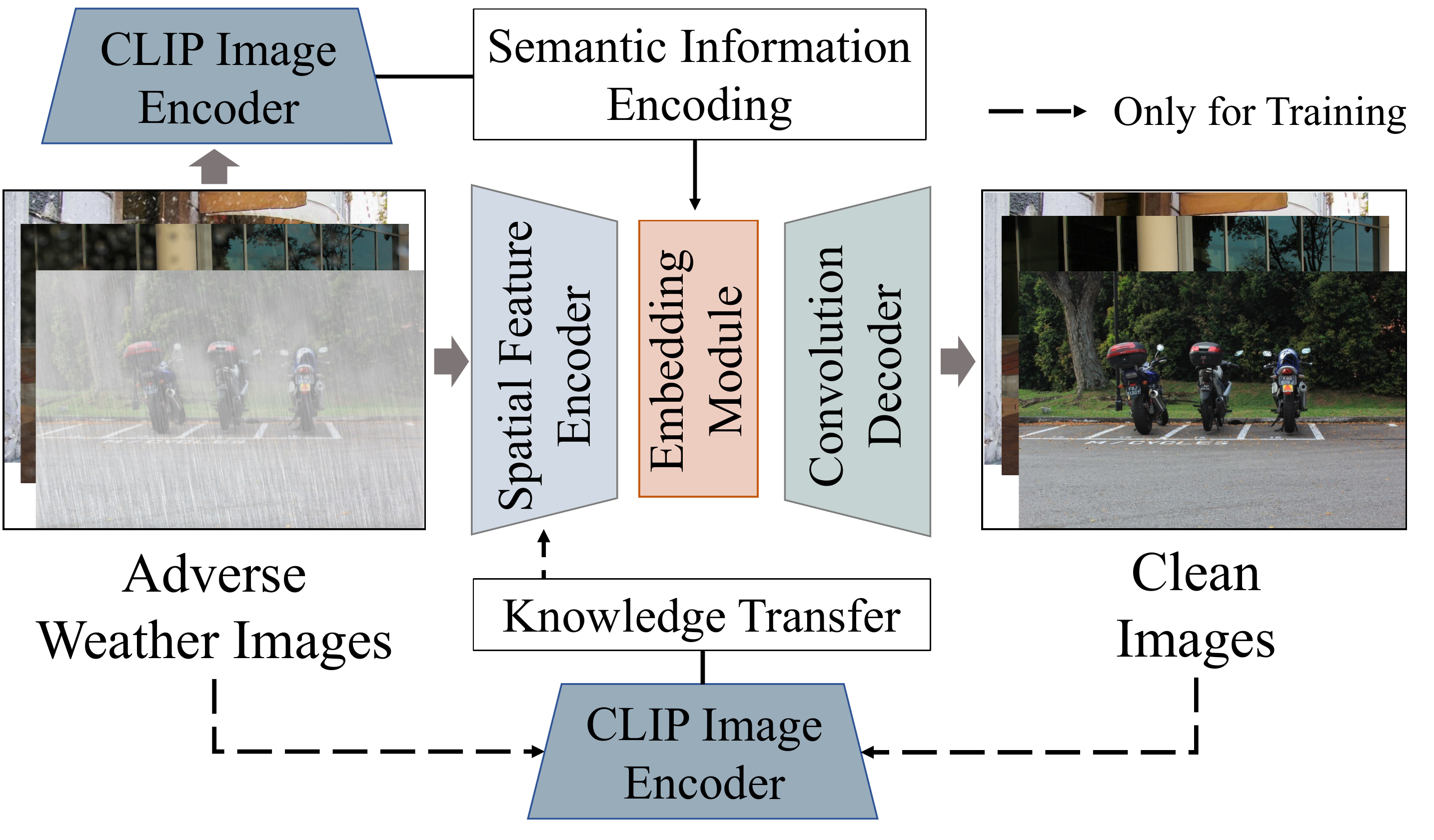}
    \caption{Overall framework of the proposed method. It tries to take advantage of CLIP image encoder from knowledge transfer for spatial feature learning and semantic information embedding.}
    \label{fig:framework}
\end{figure}

Early methods, no matter weather prior based~\cite{he2010single,roth2005fields} or deep learning based~\cite{fu2017removing,liu2018desnownet,zhao2021hybrid}, are usually designed specifically to handle one type of weather condition. Recently, removing various adverse weather in an unified framework has attracted great attention. To handle various adverse weather conditions, they either mine the spatial feature representation through encoder designs and learning~\cite{li2020all,chen2022learning}, or embed semantic information of weather type via embedding modules with learnable parameters and additional extractors~\cite{valanarasu2022transweather,li2022all}. However, almost all these methods only partially demonstrate the effectiveness of specially designed encoders and embedding modules. In this paper, we explore for better feature representation and semantic embedding simultaneously with the advance of large-scale pre-trained models (e.g., the image encoder of CLIP).

Large-scale vision or vision-language models have shown amazing semantic representation and generalization ability in various tasks. For example, CLIP~\cite{radford2021learning} has been successfully applied on various downstream tasks~\cite{patashnik2021styleclip,wei2022hairclip,gal2022stylegan,ramesh2022hierarchical,gu2021open,kamath2021mdetr,li2022language,wang2022cris,luddecke2022image,rao2022denseclip} due to its amazing semantic alignment ability between vision and language. Different from these successful practices, the multi-weather removal task depends on not only global semantic understanding but also local spatial feature representation, which is rarely explored before. Accordingly, we explore the utilization of large-scale models from both two aspects with the image encoder of CLIP in this paper, as shown in Figure~\ref{fig:framework}. \tzt{Considering that our task only involves adverse weather image inputs without text, we mainly study how to apply the image encoding capabilities of CLIP in this paper. }

\tzt{To learn distinguishable features that can effectively restore the different degraded areas in weather images, we propose a \textit{Spatially-Adaptive Residual (\textbf{SAR}) Encoder} to highlight the degraded areas, along with a \textit{Soft Residual Distillation (\textbf{CLIP-SRD})} strategy to guide its training. Accurately locating degraded areas can benefit image restoration, so we regard the residual feature from CLIP image encoder between adverse weather and clean images as the teacher knowledge and transfer them into the proposed \textbf{SAR} encoder during training. Specifically, in terms of architecture design, the proposed SAR encoder consists of a stack of SAR Transformer blocks, which can adaptively extract the potential degraded areas through spatially-adaptive residual convolutions. In terms of training strategy, due to the different optimization objectives between CLIP and our weather removal task, we carefully design a soft feature matching mechanism to transfer the knowledge from the intermediate features of CLIP image encoder into the corresponding blocks of SAR encoder.}

Semantic information such as weather type prior facilitates the model to handle different weather conditions adaptively. Thus, we propose a \textit{CLIP Weather Prior (\textbf{CWP}) Embedding Module} to take full advantage of the sample specific prior information extracted by CLIP image encoder and also the distribution specific prior learned during training. Regarding them as the key and value, as well as features from the SAR encoder as query, we embed weather type prior through the cross attention mechanism. We also design a text-based cross-entropy classification loss to supervise the training.

Following the previous works~\cite{li2020all,valanarasu2022transweather,ozdenizci2023restoring}, we conduct extensive experiments on various adverse weather removal datasets, including heavy rain~\cite{li2019heavy}, rain drop~\cite{qian2018attentive} and snow~\cite{liu2018desnownet}. We also discuss some simple and effective ways covering data augmentation and optimization objectives to improve the performance. Experimental results demonstrate the superior performance of our proposed method on multi-weather removal task. \tzt{Experiments indicate that CLIP can enhance the performance of adverse weather removal through both local spatial representation and global semantic understanding. Furthermore, we observe that the CLIP weather prior contributes more significantly to this improvement than CLIP knowledge transfer. This could be attributed to the fact that the default training objective of CLIP is to align global semantics between images and text, rather than spatial representation.
}

\section{Related Work}
\label{sec:related}
\subsection{Adverse Weather Removal}
Adverse weather removal has been extensively explored in the literature, including deraining~\cite{kang2011automatic,fu2017removing,li2019heavy,wang2020model,yang2019joint,yasarla2020exploring,zhu2017joint}, desnow~\cite{liu2018desnownet,ren2017video,chen2020jstasr,zhang2021deep,chen2021all}, dehazing~\cite{dong2020multi,li2018single,berman2016non,cai2016dehazenet,ren2016single} and rain drop removal~\cite{qian2018attentive,quan2021removing,quan2019deep,you2015adherent,liu2019dual,xiao2022image}.

\noindent\textbf{Single Weather Removal:} Different types of weather phenomenons have always been modelled theoretically in literature. 
For example, rain drop~\cite{qian2018attentive} is always modelled as: $I_{weather}=(1-M)\odot I_{clean} + R$, where $M$ is the mask and $R$ is the rain drop residual map. 
Heavy rain effected by rain streaks and haze~\cite{li2019heavy} is modelled as: $I_{weather} = T \odot (I_{clean} + \sum_{i}^{n}{R_{i}}) + (1 - T) \odot A$, where $T$ is the transmission map, $R_{i}$ is the rain streaks at the $i$-th layer and $A$ is the global atmospheric light of the scene. 
Snow~\cite{liu2018desnownet} is generally modelled as: $I_{weather} = (1 - M) \odot I_{clean} + M \odot S$, where $S$ is the snow flakes. 
Thus, most methods focus on the restoration of specific adverse weather based on these physical formulations. For rain streak removal, Deep learning based methods make great progress through recurrent network~\cite{yang2019joint}, conditional GANs~\cite{zhang2021deep}, spatial attention~\cite{wang2019spatial}, or conditional VAEs~\cite{du2020conditional}. For dehazing, attention based network~\cite{liu2019griddehazenet}, GANs~\cite{yang2018towards} and dense network~\cite{zhang2021hierarchical,zhang2018densely} are widely adopted. These technologies are also explored in rain drop removal~\cite{qian2018attentive,quan2019deep} and desnowing~\cite{liu2018desnownet,li2019stacked,zhang2021hierarchical,chen2020jstasr}. Although excellent performance has been achieved for specific weather removal, these methods can not be directly applied to other types of adverse weather removal tasks.

\noindent\textbf{Multiple Weather Removal:} Recently, restoring images under multiple degradation with single deep learning framework~\cite{liang2021swinir,chen2021pre,wang2022uformer,zamir2022restormer} has attracted growing attention. As for weather removal task, Zou et al.~\cite{zou2020deep} propose an unified framework to separate superimposed images. Zamir et al~\cite{zamir2021multi} propose a multi-stage image restoration strategy to refine the features during image restoration process. Although these methods support various weather types within an unified network, they require individual pre-trained weights for each condition. Later, Li et al.~\cite{li2020all} propose an All-in-one framework which consists of multiple task-specific encoders and a common decoder to handle various tasks under an unified framework and pre-trained weights. Valanarasu et al.~\cite{valanarasu2022transweather} build a Transformer-based~\cite{vaswani2017attention} framework and learn a set of weather queries during training to understand and adjust the weather degradation type in the image. Chen et al.~\cite{chen2022learning} propose to transfer the knowledge from several well pre-trained teacher models to one student network through a two-stage knowledge learning mechanism. $\ddot{O}$zdenizci et al.~\cite{ozdenizci2023restoring} model the process of image restoration through denoising diffusion models~\cite{ho2020denoising,song2020denoising} and design a patch-based sampling strategy for arbitrary sized image processing. Different from these methods which only focus on either spatial feature representation or semantic embedding, we integrate them into an unified structure with the advance of existing large-scale models.

\begin{figure*}[t]
    \centering
    \includegraphics[width=0.95\linewidth]{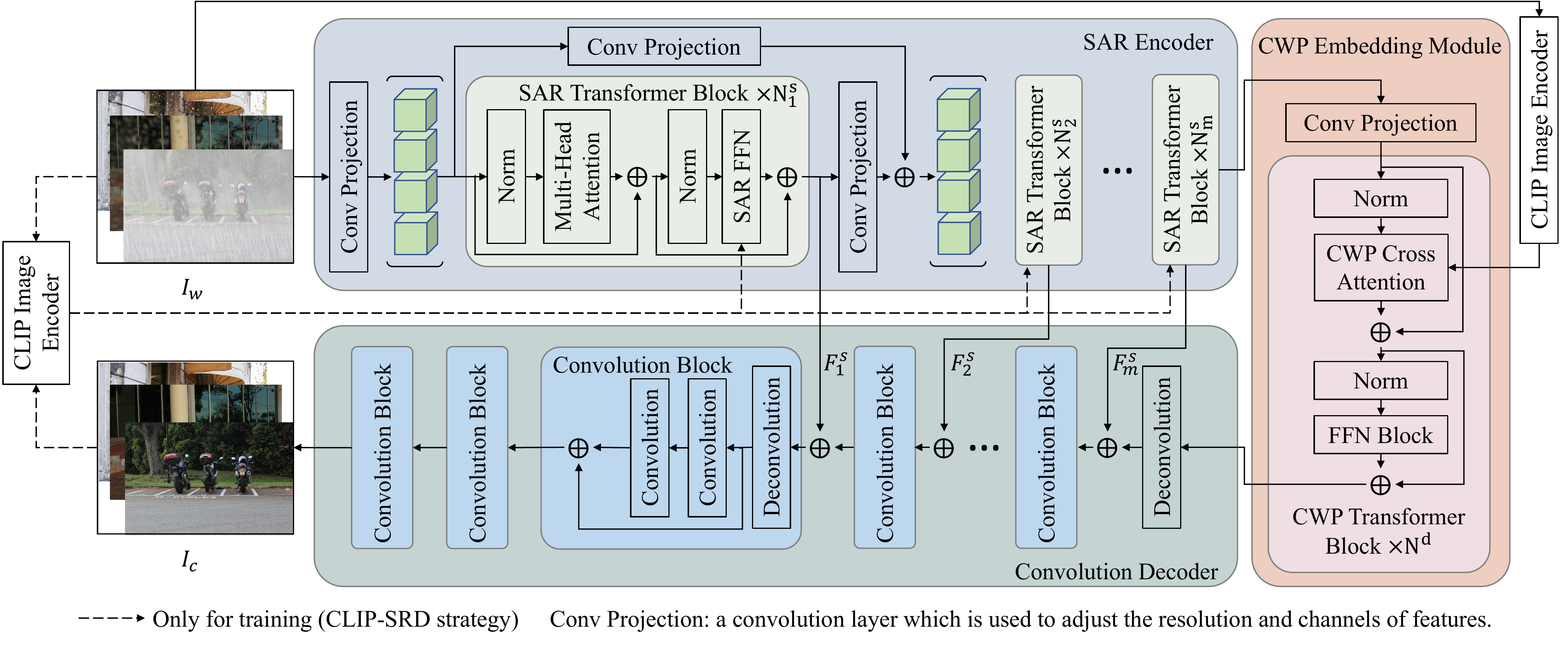}
    \caption{Architecture of the proposed method. It consists of a \textit{Spatially-Adaptive Residual Encoder} (SAR encoder), a \textit{CLIP Weather Prior Embedding Module} (CWP embedding module) and a convolution decoder. CLIP SRD strategy is only used during training.}
    \label{fig:network}
\end{figure*}

\subsection{Applications of Large-scale Models}
Recently, large-scale models, especially vision-language pre-trained models have been explored to facilitate various downstream tasks~\cite{lei2021less,lu2019vilbert,su2019vl,jia2021scaling,yao2021filip,wang2021simvlm,mustafa2022multimodal,radford2021learning}. Patashnik et al.~\cite{patashnik2021styleclip} propose a text-driven image  manipulation method based on the powerful representation of CLIP~\cite{radford2021learning}. Wei et al.~\cite{wei2022hairclip} use CLIP to guide the hairstyle generation. Gu et al.~\cite{gu2021open} and Kamath et al.~\cite{kamath2021mdetr} combine CLIP with object detection to find the specified objects through texts. Further more, CLIP is used for dense prediction tasks like segmentation~\cite{li2022language,wang2022cris,luddecke2022image,rao2022denseclip}. These works demonstrate that CLIP can be applied to image generation tasks or high-level image understanding tasks. However, there is no successful experience showing that it can be applied to low-level image processing tasks. Here, we mainly explore the potential application of CLIP image encoder in weather removal task.

\subsection{Knowledge Distillation}
Knowledge distillation transfers the knowledge of a  teacher model to a  student network to improve the performance of student network~\cite{hinton2015distilling}. It is adopted in image classification firstly by using the class probabilities produced from the teacher model as soft targets to train the student network~\cite{ba2014deep,urban2016deep}. Later, transferring the intermediate features rather than only the final predictions has been proposed by~\cite{wang2018progressive} and further improves the effect of knowledge distillation. This idea is also applied to various tasks~\cite{wang2019distilling,liu2019structured,li2017mimicking,xie2018improving,chen2022learning}. However, the knowledge from CLIP image encoder may not be suitable for direct distillation since it is not optimized for the same objective as the student network. Hence, we propose the soft residual matching mechanism to transfer spatial degradation information from CLIP.

\section{Method}
Given an image under different adverse weather conditions $I_{w}\in\mathbb{R}^{H\times W}$, we aim to restore its corresponding clean version $I_{c}$. Our proposed method follows the similar \yw{encoder-decoder} framework in~\cite{valanarasu2022transweather}. 
As shown in Figure~\ref{fig:network}, SAR encoder extracts \yw{degradation spatially-adaptive features} from input images and is trained via CLIP-SRD strategy. 
CWP embedding module embeds the weather type prior into features. Finally, convolution decoder takes the output of CWP embedding module and progressively restores clean images with the help of four multi-scale intermediate features from SAR encoder. 
To take advantage of CLIP, we transfer the spatial representation of CLIP to SAR encoder during training and embed the weather prior from CLIP to CWP embedding module.
\tzt{In the subsequent sections, we will introduce the architecture of SAR encoder (\Sref{sec:sarffn}), the CLIP-SRD strategy (\Sref{sec:clipsrd}), CWP embedding module (\Sref{sec:cwp}) and the training of the whole network (\Sref{sec:training}). }

\subsection{Spatially-Adaptive Residual Encoder}
\label{sec:sarffn}
Encoder plays an important role for various visual tasks to project original discrete RGB signal to high-dimensional feature space while filtering out interference information. To enhance its ability, most of recent methods focus on well designed network modules and training strategy. For architecture design, All-in-one~\cite{li2020all} searches specific encoder architecture for each weather type. TransWeather~\cite{valanarasu2022transweather} proposes Intra-Patch Transformer Block to extract more details. As for training strategy, Chen et al.~\cite{chen2022learning} propose to transfer the knowledge from multiple teacher models into student network during training, while each teacher model is trained for a single weather type. To enable the encoder to distinguish different degraded areas, we propose the \textit{Spatially-Adaptive Residual Transformer Block} which extracts residual features through spatially-adaptive convolutions, and guide its training with the large-scale pre-trained CLIP image encoder (see \Sref{sec:clipsrd}).

Figure~\ref{fig:network} shows the detailed architecture of our SAR encoder. It is stacked with $m$ hierarchical stages ($m=4$). We build skip connection between two adjacent stages. Each stage consists of several SAR Transformer blocks. A single convolution layer (defined as \textit{Conv Projection}) is used to reduce the resolution and increase the number of channels of features. The SAR Transformer block follows the classical architecture~\cite{vaswani2017attention} with a specially designed \textit{SARFFN}, which can be formulated as:
\begin{equation}
    \begin{split}
        F_i^{'} = MSA(Norm(F_{i-1})) + F_{i-1}, \\
        F_i = SARFFN(Norm(F_i^{'})) + F_i^{'}, \\
    \end{split}
    \label{eq:block}
\end{equation}
where $F_{i-1}$ and $F_i$ denote the input and output of the $i$-th block. $MSA$ is the multi-head self attention. Inspired by~\cite{valanarasu2022transweather,wang2021pyramid,wu2021cvt}, we use the efficient self-attention mechanism which reduces the resolution of key and value through convolution. 

\begin{figure}[t]
    \centering
    \includegraphics[width=0.8\linewidth]{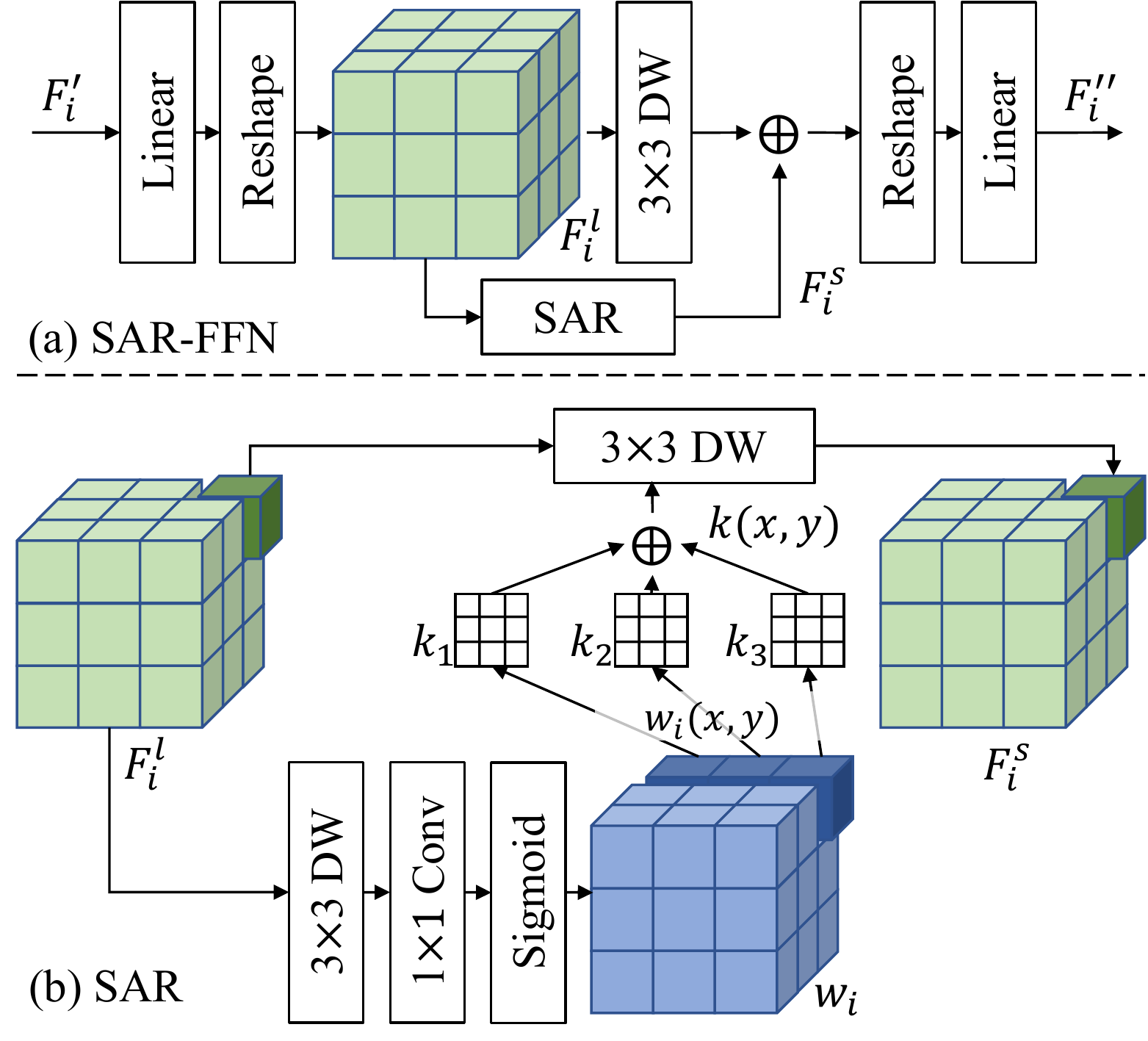}
    \caption{(a) Architecture of the proposed \textit{SARFFN}. $DW$ is the Depth-Wise Separable Convolution~\cite{chollet2017xception}. (b) Details of \textit{Spatially-Adaptive Residual (SAR)}. Activation layers are ignored for convenience.}
    \label{fig:sarffn}
\end{figure}

\yw{\textit{SARFFN} is the \textit{Spatially-Adaptive Residual FFN} which can extract degraded areas adaptively. 
We show the details of \textit{SARFFN} in Figure~\ref{fig:sarffn}.} Inspired by recent work~\cite{li2021localvit,xie2021segformer,valanarasu2022transweather}, we use convolutions to introduce local and positional information into Transformers. Let $F_{i}^{'}$ and $F_{i}^{''}$ define the input features and the output features, the overall operation can be summarized as:
\begin{equation}
    \centering
        F_{i}^{l} = Reshape(Linear(F_{i}^{'})),
\end{equation}
\begin{equation}
    \centering
        F_{i}^{''} = Linear(Reshape(DW(F_{i}^{l})+SAR(F_{i}^{l}))),
\end{equation}
where $SAR$ is the proposed \textit{Spatially-Adaptive Residual} module, $DW$ is the Depth-Wise Separable Convolution, $F_{i}^{l}$ is the input of $SAR$ module, $Reshape$ changes the shape of features to match the subsequent operations.

Considering the spatial variations of degraded areas, \yw{we specially design a spatially-adaptive dynamic convolution to extract residual features in our $SAR$ module.} Different from the normal convolution which freezes the filters after training, dynamic convolutions~\cite{jia2016dynamic,chen2021dynamic,yang2019condconv} generate dynamic weights conditioned on the input, and have been applied to many visual tasks~\cite{xu2020unified,hu2019meta,liu2019learning,park2019semantic,wang2021image,tian2020conditional}. Among them, Condconv~\cite{yang2019condconv} produces dynamic weights by combining a set of normal convolution filters conditionally, which makes the inference efficient. Here, we extend it to support spatially-adaptive modulation and apply Depth-wise filters~\cite{chollet2017xception} for efficiency (similar design is also used in~\cite{tian2020conditional}).
As shown in Figure~\ref{fig:sarffn} (b), we define a $3\times3$ Depth-Wise Separable Convolution with $N_{dw}$ kernels, and predict the weights of these kernels in each location through several efficient convolutions. Let $w_{i}\in\mathbb{R}^{N_{dw}\times H_{i}\times W_{i}}$ define the weights map for $F_{i}^{l}\in\mathbb{R}^{C_{i}\times H_{i}\times W_{i}}$, where $N_{dw}$ is the pre-defined number of kernels, $H_{i}$ and $W_{i}$ are the height and width of current features, and $C_{i}$ is the number of channels. $w_{i}$ can be obtained as follows:
\begin{equation}
    w_{i} = Sigmoid(Conv(DW(F_{i}^{l}))).
\end{equation}
In Figure~\ref{fig:sarffn} (b), we set $N_{dw}$ to 3 as an example. Then, the spatially-adaptive kernel at location $(x,y)$ is calculated as:
\begin{equation}
    k_{i}(x,y) = \sum_{j=1}^{N_{dw}} w_{i,j}(x,y) \otimes k_{i,j},
\end{equation}
where $\otimes$ is the element-wise multiplication. Finally, we use $k_{i}(x,y)$ to calculate the values at location $(x,y)$ in residual features $F_{i}^{s}$.

\subsection{CLIP Soft Residual Distillation}
\label{sec:clipsrd}
To guide the learning of $F_{i}^{s}$, we propose to transfer the knowledge from CLIP during training. Since the original CLIP is not trained for adverse weather removal, we do not directly use the CLIP features extracted from weather images $I_{w}$. Instead, we calculate the residual features between weather images $I_{w}$ and clean images $I_{c}$. Specifically, let $F_{j}^{C}(I_{w})$ and $F_{j}^{C}(I_{c})$ be the features of the $j$-th stage in CLIP image encoder for weather images and clean images respectively. we propose to use \yw{the soft feature matching mechanism with} normalized L1 distance to transfer the knowledge to all residual maps in the $j$-th stage:
\begin{equation}
    \mathcal{L}_{j}^{d} = \sum_{i=1}^{N_{j}^{s}} ||norm(F_{i,j}^{s})-norm(F_{j}^{C}(I_{c})-F_{j}^{C}(I_{w}))||_{1},
\end{equation}
where $norm(\cdot)$ is the normalization operation, $N_{j}^{s}$ is the number of Transformer blocks in the $j$-th stage. We use a channel-wise adaptive pooling layer to match the channels of these features if necessary. The final distillation loss $\mathcal{L}^{d}$ is as follows:
\begin{equation}
    \mathcal{L}^{d} = \frac{1}{mN_{j}^{s}} \sum_{j=1}^{m} \mathcal{L}_{j}^{d}.
\end{equation}

\subsection{CLIP Weather Prior Embedding Module}
\label{sec:cwp}
To process various weather types in an unified way, some methods~\cite{valanarasu2022transweather,wang2022uformer} learn a set of  parameters to adapt to weather. However, these parameters only learn weather distributions in training datasets which cannot distinguish the exact weather type of images. AirNet~\cite{li2022all} designs an encoder to extract the degradation prior directly, but needs to be trained carefully through contrastive learning. Differently, we propose to use the weather prior from well pre-trained CLIP image encoder together with distribution-based weather prior from a set of learnable parameters.

As shown in Figure~\ref{fig:network}, the proposed \textit{CLIP Weather Prior Embedding Module} consists of $N^{d}$ CWP Transformer blocks, which have the similar architecture with that of SAR Transformer blocks. The differences are two fold: 1)  $FFN$ is the classical version without $SAR$ architecture;  2) the multi-head attention is replaced by the proposed CWP cross attention to modulate features adaptively.

\begin{figure}[t]
    \centering
    \includegraphics[width=0.9\linewidth]{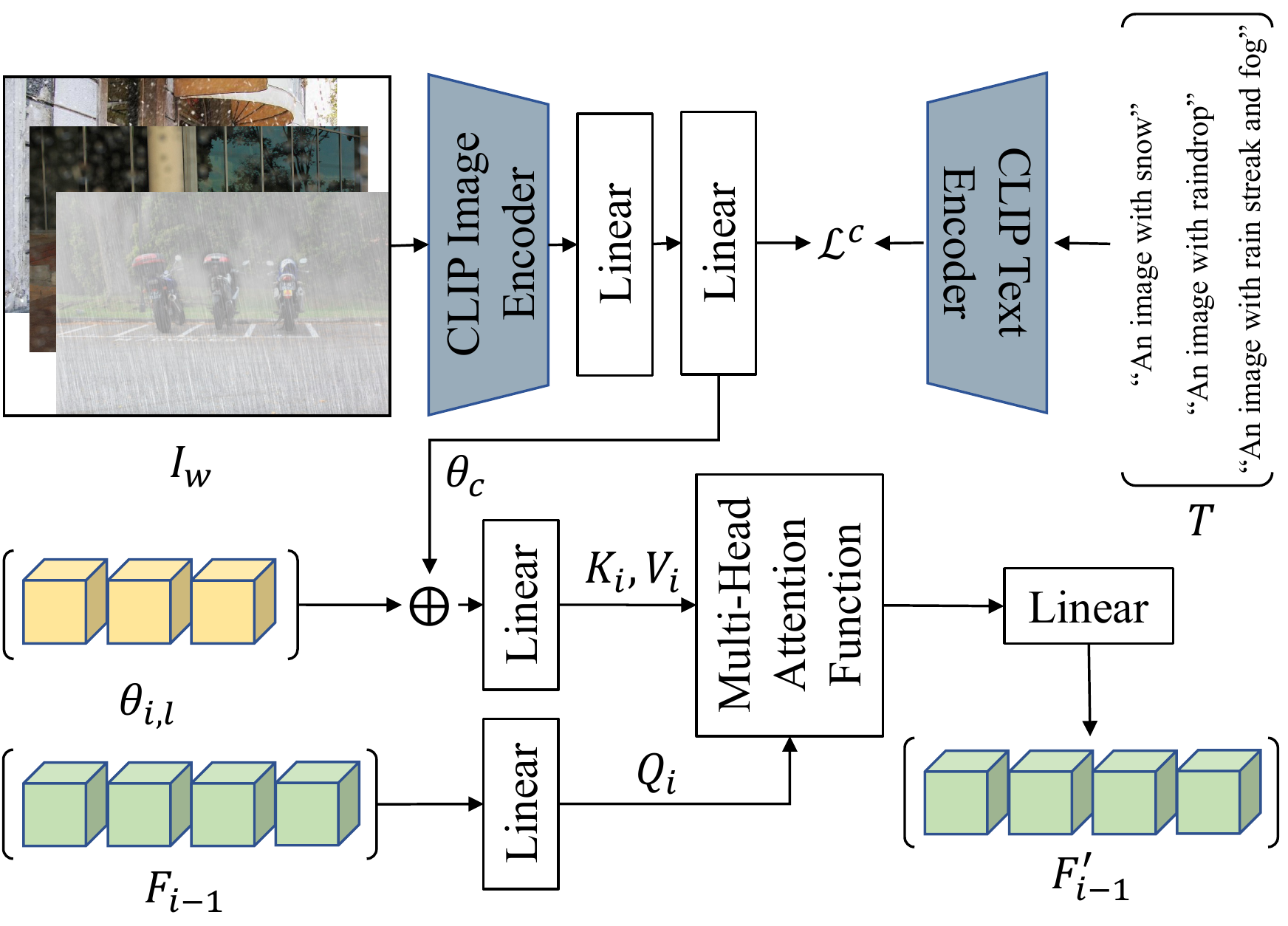}
    \caption{Illustration of the proposed CWP cross attention. Activation layers are ignored for convenience. CLIP text encoder is only used during training.}
    \label{fig:dwpca}
\end{figure}

Figure~\ref{fig:dwpca} shows the cross attention mechanism proposed to embed dual weather prior. Specifically, we obtain the input specific weather prior embedding $\theta_{c}$ via CLIP image encoder along with two linear layers. let $\theta_{i,l}\in\mathbb{R}^{L_{\theta}\times C_{i}}$ denote the learnable weather prior embedding in the $i$-th CWP cross attention, where $L_{\theta}=48$ is the pre-defined hyper-parameter. The key and value for multi-head attention are calculated as:
\begin{equation}
    K_{i}, V_{i} = Linear(\theta_{i,l}+w_{i,c}\theta_{c}),
\end{equation}
where $w_{i,c}$ is a learnable weight which is initialized as $0$. Following the definition of equation~\ref{eq:block}, let $F_{i-1}$ and $F_{i-1}^{'}$ be the input and output features in the $i$-th CWP cross attention respectively. We use the feature tokens from $F_{i-1}$ as query $Q_{i}=Linear(F_{i-1})$ to fuse with the key and value.
After that, we can get the output $F_{i-1}^{'}$ through multi-head cross attention which is similar with the original version~\cite{vaswani2017attention} along with a linear layer: 
\begin{equation}
    F_{i-1}^{'} = Linear(softmax(\frac{Q_{i}K_{i}^{T}}{\sqrt{C_{i}}}) V_{i}).
\end{equation}

To match the number of channels of original CLIP image encoder output with $C_{i}$ and project it to the same feature space as $\theta_{i,l}$, we add two linear layers at the end of CLIP image encoder and train them along with the whole network. In addition, we propose a text-based classification loss $\mathcal{L}^{c}$ to facilitate their training. According to the weather types we want to handle, we firstly design the text prompting $T$ for each weather type and use the pre-trained CLIP text encoder to get the corresponding feature representations $F_{T}$. Then, we calculate the cosine distance between the text features and image features, and get the similarity tensor that can be regarded as the potential probability of which weather each image belongs to. Finally, a Cross Entropy function is used to get loss $\mathcal{L}^{c}$ for training. Sine the text selection is not the focus in this paper, we simply use the prompt template of ``An image with XXX'', where ``XXX'' is the type of adverse weather.

\subsection{Training Strategy}
\label{sec:training}
Except the proposed modules mentioned before, we also explore some simple and effective ways to train our network, including data augmentation and loss function.

\noindent\textbf{Data Augmentation.} To ensure the consistent performance of various weather removal, we use Cut-Mix~\cite{yun2019cutmix} to combine arbitrary adverse weather images, which allows the network learn to restore these types of degradation at the same time. The ratio of it is defined as $\rho_{mix}$.
Other data augmentation ways are explored in supplementary material. Notely, the text features are only used to calculate the similarity of classification. Thus, we only need to change the original one-hot label into the soft one when using Cut-Mix.

\noindent\textbf{Loss Function.} Following the previous work~\cite{valanarasu2022transweather}, we use smooth L1 loss $\mathcal{L}^{s}$ and perceptual loss~\cite{johnson2016perceptual} $\mathcal{L}^{p}$ to train our proposed method. In addition, we use structural similarity (SSIM) loss $\mathcal{L}^{S}$ and Peak Signal to Noise Ratio (PSNR) loss $\mathcal{L}^{P}$ which are calculated as follows:
\begin{equation}
    \begin{split}
        \mathcal{L}^{S} = 1 - ssim(I_{o}, I_{c}),\\
        \mathcal{L}^{P} = 1 - psnr(I_{o}, I_{c}) / 100,\\
    \end{split}
\end{equation}
where $I_{o}$ is the output of network, $ssim(\cdot)$ and $psnr(\cdot)$ are the functions to calculate SSIM and PSNR scores. 

The final loss function $\mathcal{L}$ is as follows:
\begin{equation}
    \mathcal{L} = 
    \mathcal{L}^{s}+\lambda_{p} \mathcal{L}^{p}+\lambda_{S} \mathcal{L}^{S}+\lambda_{P} \mathcal{L}^{P}+\lambda_{c} \mathcal{L}^{c}+\lambda_{d} \mathcal{L}^{d},
\end{equation}
where $\lambda_{p}$, $\lambda_{S}$, $\lambda_{P}$, $\lambda_{c}$ and $\lambda_{d}$ are the weights of corresponding losses. Please note that we add knowledge distillation loss $\mathcal{L}^{d}$ after training $N_{d}$ epochs.

\begin{table*}[t]
  \centering
  \caption{Comparison with SOTA weather specific methods and multi-weather removal methods. $^\dagger$ means the scores are provided by the paper. $^*$ means the method without additional PSNR/SSIM loss and data augmentation. Weather specific methods are only trained on their own weather dataset (the corresponding results are marked in \textit{italics}), and are also directly tested on other weather removal sets if possible. Multi-weather methods are trained on the same multi-weather datasets mentioned in Section\ref{sec:datasets}.}
  \begin{tabular}{c|c|c|c|c|c|c|c|c|c}
    \toprule
    Weather & \multirow{2}{*}{Methods} & \multicolumn{2}{c|}{Snow100K-L~\cite{liu2018desnownet}} & \multicolumn{2}{c|}{Test1~\cite{li2019heavy}} & 
    \multicolumn{2}{c|}{RainDrop~\cite{qian2018attentive}} & \multicolumn{2}{c}{Mean}\\
    \cline{3-10}
    Type &  & PSNR$\uparrow$ & SSIM$\uparrow$ & PSNR$\uparrow$ & SSIM$\uparrow$ & PSNR$\uparrow$ & SSIM$\uparrow$ & PSNR$\uparrow$ & SSIM$\uparrow$\\
    \midrule
    \multirow{2}{*}{Snow} & DesnowNet$^\dagger$~\cite{liu2018desnownet} & \textit{27.17} & \textit{0.8983} & - & - & - & - & - & -\\
    & DDMSNet$^\dagger$~\cite{zhang2021deep} & \textit{28.85} & \textit{0.8772} & - & - & - & - & - & -\\
    \midrule
    rain streak & HRGAN$^\dagger$~\cite{li2019heavy} & - & - & \textit{21.56} & \textit{0.8550} & - & - & - & - \\
    \&. haze & MPRNet~\cite{zamir2021multi} & 18.99 & 0.7320 & \textit{28.49} & \textit{0.9337} & 19.43 & 0.8689 & 22.30 & 0.8448\\
    \midrule
    \multirow{2}{*}{Raindrop} & AttentiveGAN~\cite{qian2018attentive} & 18.36 & 0.6659 & 14.05 & 0.6469 & \textit{31.47} & \textit{0.9582} & 21.29 & 0.7570\\
    & CCN$^\dagger$~\cite{quan2021removing} & - & - & - & - & \textit{31.34} & \textit{0.9500} & - & -\\
    \midrule
    \midrule
    \multirow{3}{*}{Multi-weather:} & MPRNet~\cite{zamir2021multi} & 27.92 & 0.9108 & 28.08 & 0.9306 & 29.45 & 0.9416 & 28.48 & 0.9277\\
    & Restormer~\cite{zamir2022restormer} & 27.76 & 0.9067 & 27.24 & 0.9206 & 29.29 & 0.9371 & 28.10 & 0.9214 \\
    & Uformer~\cite{wang2022uformer} & 26.60 & 0.8872 & 25.40 & 0.8886 & 27.38 & 0.9193 & 26.46 & 0.8984 \\
    Snow \&.  & All-in-one$^\dagger$~\cite{li2020all} & 28.33 & 0.8820 & 24.71 & 0.8980 & \textbf{31.12} & 0.9288 & 28.05 & 0.9029\\
    Rain streak& TransWeather~\cite{valanarasu2022transweather} & 28.14 & 0.9136 & 27.64 & 0.9285 & 29.53 & 0.9466 & 28.44 & 0.9296 \\
    \&. Haze \&. & Chen et al.~\cite{chen2022learning} & 26.96 & 0.8967 & 24.20 & 0.9037 & 30.47 & 0.9541 & 27.21 & 0.9182\\
    Raindrop & Weatherdiff~\cite{ozdenizci2023restoring} & 27.95 & 0.9155 & 27.42 & \textbf{0.9434} & 28.62 & 0.9483 & 28.00 & 0.9357\\
    \cline{2-10}
    & Ours &  \textbf{28.54} & \textbf{0.9221} & 28.68 & 0.9400 & 30.40 & \textbf{0.9558} & \textbf{29.21} & \textbf{0.9393}\\
    & Ours$^*$ & 28.45 & 0.9178 & \textbf{28.70} & 0.9362 & 29.98 & 0.9491 & 29.04 & 0.9343 \\
    \bottomrule
  \end{tabular}
  \label{tab:comp_sota}
\end{table*}

\begin{table}[t]
    \centering
    \caption{Quantitative comparison of main designs. }
    \begin{tabular}{c|l|c|c}
    \toprule
    & Methods & PSNR $\uparrow$ & SSIM $\uparrow$ \\
    \midrule
    & Baseline~\cite{valanarasu2022transweather} & 28.89 & 0.9458 \\
    \midrule
    A &+ PSNR and SSIM Loss & 29.04 & 0.9498 \\
    \midrule
    B &A + Data Augmentation & 29.23 & 0.9514 \\
    \midrule
    C &B + CWP Embedding Module & 29.43 & 0.9523 \\
    \midrule
    D &C + SAR Encoder and CLIP SRD & \textbf{29.71} & \textbf{0.9544} \\
    \bottomrule
    \end{tabular}
    \label{tab:ablation}
    \vspace{-0cm}
\end{table}

\section{Experiments}
\subsection{Implementation Details}
We train our network via Adam Optimizer~\cite{kingma2014adam} ($\beta_1=0.9$, $\beta_2=0.999$) for total $250$ epochs with a batch size of $32$. The learning rate is set to $2e^{-4}$ and is reduced by half after every $100$ epochs. We empirically set $\lambda_{p}$, $\lambda_{S}$, $\lambda_{P}$, $\lambda_{c}$ and $\lambda_{d}$ to $0.04$, $0.1$, $0.02$, $0.08$ and $0.1$. $N_{d}$ is set to $200$. We set $\rho_{mix}$ to $0.7$ through experiments. We use the class token of ViT-B/32 based CLIP image encoder ($CLIP_{V}$) as the weather prior in CWP embedding module, and the intermediate features of ResNet50x4 based CLIP image encoder ($CLIP_{R}$) to transfer the knowledge. Among the $5$ stages in the CLIP image encoder, we use the output of the first, third, fourth and fifth stages to transfer knowledge.
$\{N_{1}^{s}, N_{2}^{s}, N_{3}^{s}, N_{4}^{s}\}$ are set to 3,3,3,2, while $N^{d}$ is set to 3. The number of kernels $N_{dw}$ in \textit{SAR} module is set to $3$ by default. All experiments are implemented in PyTorch~\cite{paszke2019pytorch} and conducted on Tesla V100 GPUs. 

\subsection{Datasets and Metrics}
\label{sec:datasets}
Following the previous work~\cite{li2020all,valanarasu2022transweather,ozdenizci2023restoring}, we conduct experiments on three adverse weather conditions: snow, heavy rain with haze and raindrops. The training dataset consists of 18,069 images: 9,001 synthetic snow images from Snow100K~\cite{liu2018desnownet}, 818 raindrops images from RainDrop~\cite{qian2018attentive} and 8,250 heavy rain images with rain streaks and haze from Outdoor~\cite{li2019heavy} (denoted as Test1). Accordingly, the test dataset consists of 16,801 images from Snow100K-L test set~\cite{liu2018desnownet}, 58 images from RainDrop test set~\cite{qian2018attentive} and 750 images from Test1 dataset~\cite{li2019heavy}. For fair comparison, we directly use the training and test set provided by~\cite{valanarasu2022transweather}. 
The training images are randomly cropped with the resolution of $256\times 256$, while the test images keep their original resolution unchanged.
We calculate the structural similarity (SSIM) and Peak Signal to Noise Ratio (PSNR) via the code released by~\cite{valanarasu2022transweather} for quantitive comparison. When comparing our method to others, we save the results as PNG images and then calculate the metrics. Otherwise, we directly use the output of model in our ablation study for convenience.

\subsection{Comparison with SOTA Methods}
In Table~\ref{tab:comp_sota}, we compare the proposed method with recent SOTA methods, including weather specific methods and multi-weather removal methods. For weather specific methods, we select DesnowNet~\cite{liu2018desnownet} and DDMSNet~\cite{zhang2021deep} for image desnowing, HRNet~\cite{li2019heavy} and MPRNet~\cite{zamir2021multi} for image deraining and hazing, and AttentiveGAN~\cite{qian2018attentive} and CCN~\cite{quan2021removing} for removing raindrops. For multi-weather removal, we select image restoration methods (MPRNet~\cite{zamir2021multi}, Restormer~\cite{zamir2022restormer} and Uformer~\cite{wang2022uformer}) and adverse weather removal metods (All-in-one~\cite{li2020all}, TransWeather~\cite{valanarasu2022transweather}, Weatherdiff~\cite{ozdenizci2023restoring} and the method proposed by~\cite{chen2022learning}). If the official training code or pre-trained models are provided, we calculate the metrics by ourselves. Otherwise we directly report the scores in their original paper.

Weather specific methods always achieve promising performance in the specific weather, sometimes even better than all of the multi-weather methods (e.g. CCN and AttentiveGAN achieve the best performance in terms of PSNR on RainDrop dataset). However, if these methods are directly applied to other weather removal tasks, their performance will degrade significantly. 
As a contrast, multi-weather removal methods can achieve consistent performance on various adverse weather removal tasks. Compared to these multi-weather removal methods, our method achieves the best performance in almost all metrics and all test datasets with the successful application of powerful CLIP. Specifically, compared to recent SOTA methods, like TransWeather and MPRNet, the proposed method improve the average PSNR and SSIM scores by more than 0.73 and 0.0097. We further provide the score that without additional PSNR/SSIM loss and data augmentation (defined as Ours$^*$) to demonstrate that it still achieves SOTA performance with the proposed two modules. We find the scores reported in All-in-one~\cite{li2020all} are unreasonably higher than that of some methods~\cite{ozdenizci2023restoring,chen2022learning} and give analysis in the supplementary material.
\begin{figure*}[t]
    \centering
    \includegraphics[width=1\linewidth]{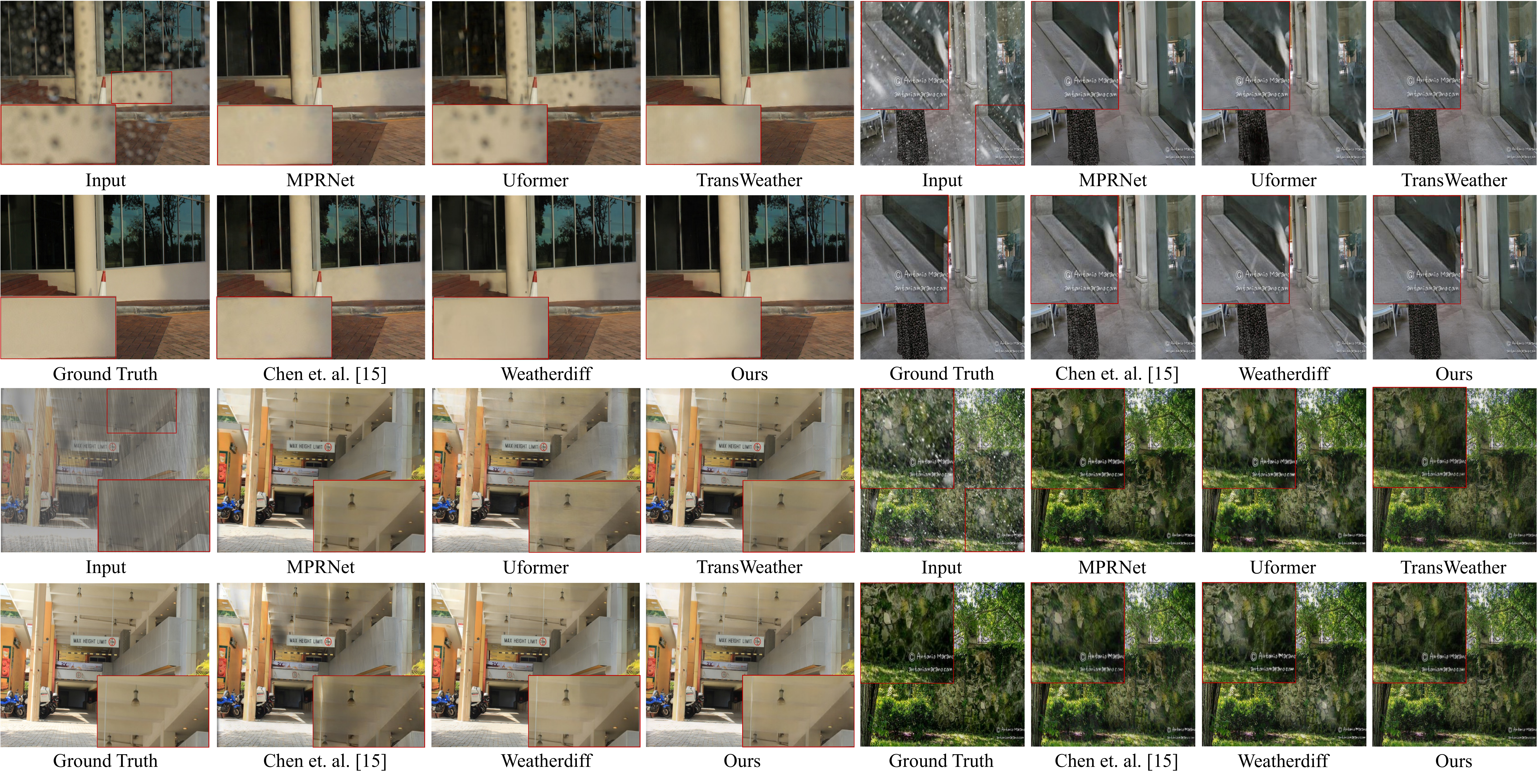}
    \caption{Visual comparison of different multi-weather methods on RainDrop~\cite{qian2018attentive},  Test1~\cite{li2019heavy} and Snow100K-L~\cite{liu2018desnownet} datasets. The patches highlighted in red box are zoom-in for better comparison.}
    \label{fig:results}
    \vspace{-0cm}
\end{figure*}

The qualitative comparisons are shown in Figure~\ref{fig:results}. Compared to other multi-weather methods, the proposed mehtod can handle various adverse weather conditions caused by raindrop, snow particles or a combination of rain streak and haze. Others can not completely restore some patches under serious degraded conditions. 

\begin{table*}[t]
    \centering
    \caption{Comparison of model parameters and FLOPs. The FLOPs of Weatherdiff~\cite{ozdenizci2023restoring} contains the number of patches (999) and the number of sampling steps (25) with the input patch size of $64\times64$. }
    \begin{tabular}{c|c|c|c|c|c}
    \toprule
    Methods & Parameters & Resolution & FLOPs & Inference Memory & Inference Times\\
    \midrule
    MPRNet~\cite{zamir2021multi} & 16M & $480\times640$ & 6.534T & 3.6G & 0.53s \\
    Uformer~\cite{wang2022uformer} & 51M & $512\times512$ & 357.8G & 2.9G & 0.33s \\
    TransWeather~\cite{valanarasu2022transweather} & 38M & $480\times640$ & 38.1G & 1.4G & 0.06s\\
    Chen et al.~\cite{chen2022learning} & 29M & $480\times640$ & 115.1G & 1.3G & 0.38s\\
    Weatherdiff~\cite{ozdenizci2023restoring} & 83M & $480\times640$ & 29.7G$\times$999$\times$25 & 10.0G & $>$1s\\
    \midrule
    Ours & 126M & $480\times640$ & 51.5G & 1.7G & 0.09s\\
    \bottomrule
    \end{tabular}
    \label{tab:efficiency}
    \vspace{-0cm}
\end{table*}

We compare the model size and computational complexity in Table~\ref{tab:efficiency}. \yw{Although the additional CLIP encoder causes the parameters of our proposed method to be larger than that of other methods, the FLOPs only increases 13.4G compared to TransWeather due to the fixed short input length (49) of CLIP image encoder and is still less than other methods. In terms of inference memory and times, our method is still efficient than most other methods, especially when compared to MPRNet and Weatherdiff.}

\subsection{\tzt{Performance on Real-world Dataset}}
\begin{table}[t]
    \centering
    \caption{Comparison on real-world dataset. All compared methods are trained on synthesized data and then directly tested on SPA-data~\cite{wang2019spatial}.}
    \begin{tabular}{c|c|c}
    \toprule
    Methods & PSNR $\uparrow$ & SSIM $\uparrow$ \\
    \midrule
    Uformer~\cite{wang2022uformer} & 28.60 & 0.9224 \\
    TransWeather~\cite{valanarasu2022transweather} & 30.65 & 0.9337 \\
    Chen et al.~\cite{chen2022learning} & 31.30 & 0.9371 \\
    Weatherdiff~\cite{ozdenizci2023restoring} & 31.43 & 0.9339 \\
    \midrule
    Ours & \textbf{31.78} & \textbf{0.9419} \\
    \bottomrule
    \end{tabular}
    \label{tab:comp_real}
\end{table}
We further compare the performance of different methods on a real-world dataset. Specifically, we test the derain performance on SPA-dataset~\cite{wang2019spatial}. All methods are only trained on the synthesized multi-weather data~\cite{valanarasu2022transweather}. As shown in Table~\ref{tab:comp_real}, our proposed method still achieves the best performance. Notely, all the methods are not trained on the rain conditions (heavy rain data contains rain and haze). It may result in the scores of methods being lower than some derain methods reported in~\cite{wang2019spatial}. 

\begin{figure*}
    \centering
    \includegraphics[width=1\linewidth]{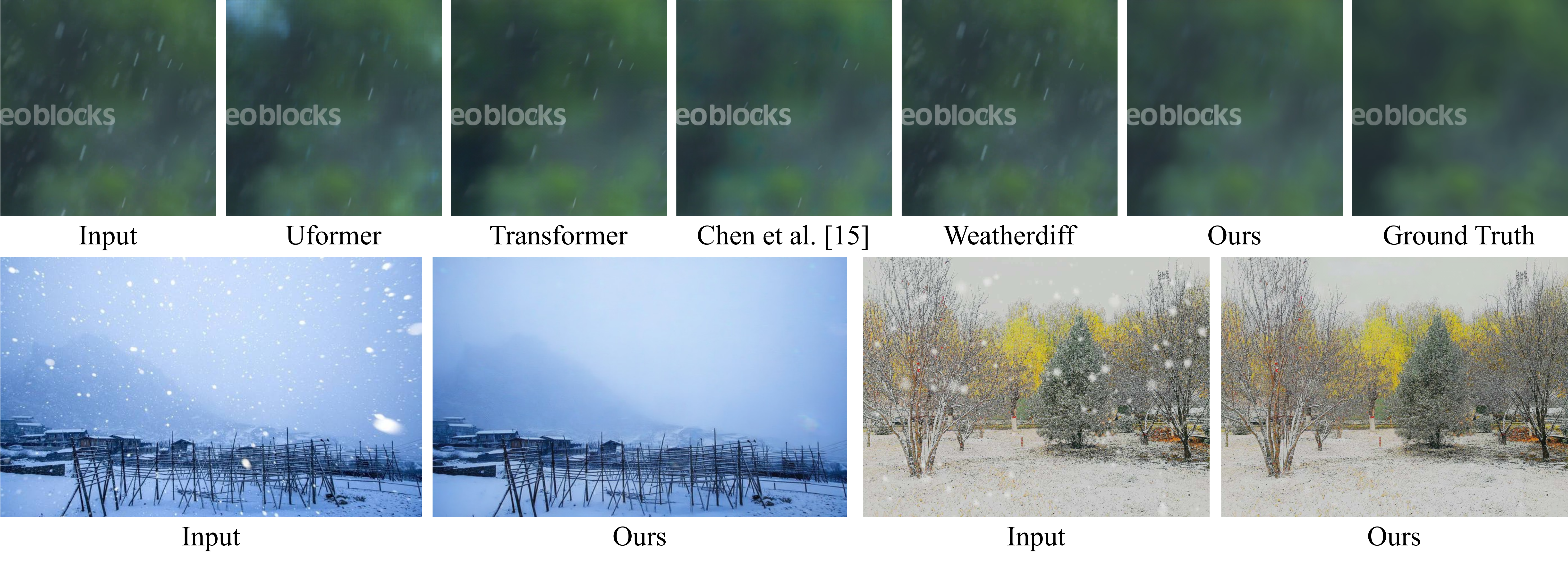}
    \caption{Top: visual comparison of different methods for real-world rain images (SPA-dataset~\cite{wang2019spatial}); Bottom: visual results of our proposed method for real-world snow images.}
    \label{fig:vis_real}
\end{figure*}

We show the derain qualitative comparisons on the top of Figure~\ref{fig:vis_real}. Although almost all methods can remove rain streaks to a certain extent, our proposed method performs the best. On the bottom of Figure~\ref{fig:vis_real}, we also show some real-world snow  removal results which comes from the Internet to demonstrate the performance of our proposed method. 

\subsection{Main Ablation Study}
We conduct extensive ablation experiments to demonstrate the effectiveness of each proposed module and to find the appropriate hyper-parameters. 

\begin{figure*}
    \centering
    \includegraphics[width=0.8\linewidth]{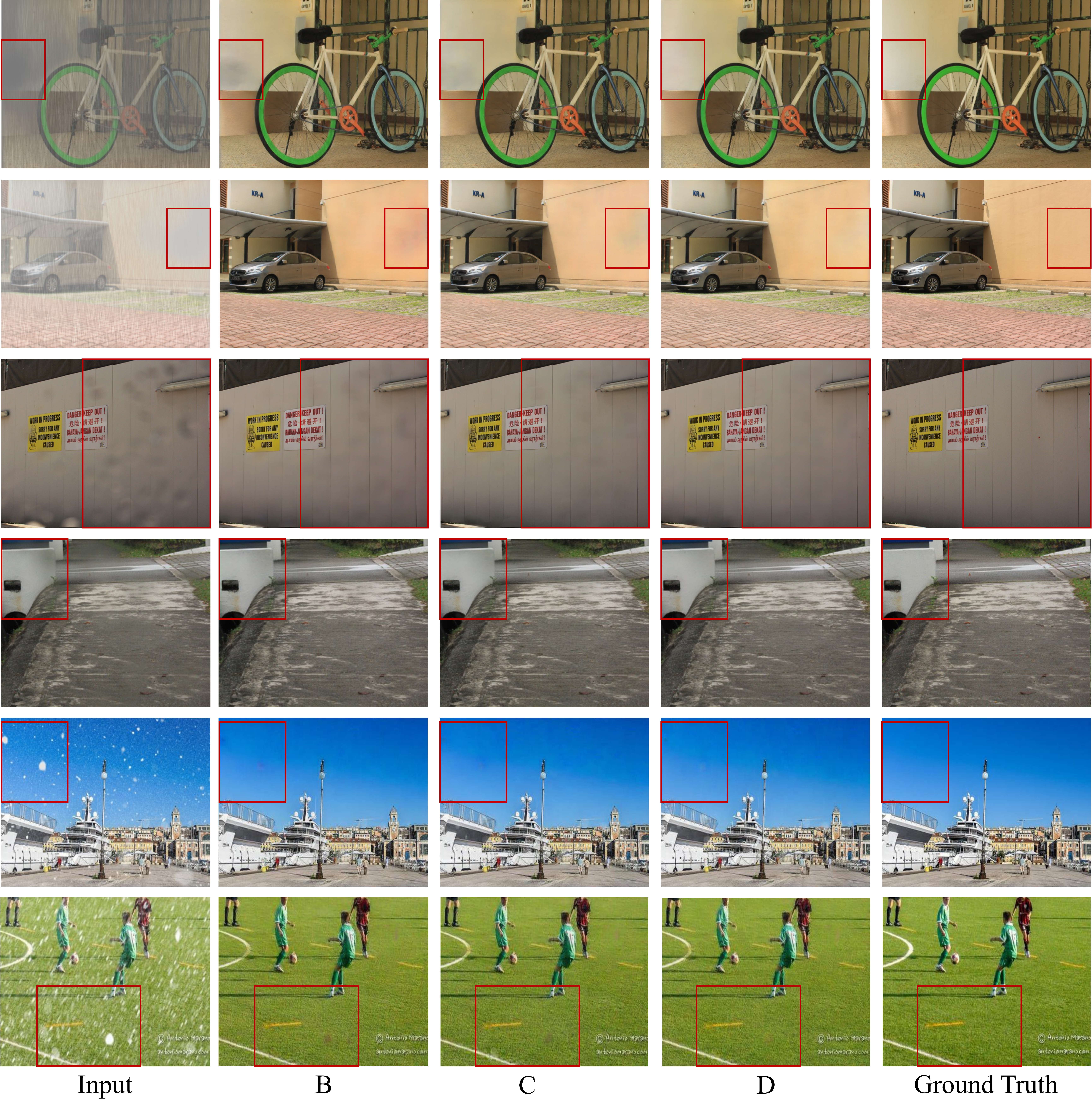}
    \caption{Visual comparison of different versions under heavy rain with haze, raindrop and snow weather conditions. $B, C, D$ are defined in Table~\ref{tab:ablation}.}
    \label{fig:vis_bcd}
\end{figure*}

\noindent\textbf{Effectiveness of Main Designs.} We first summarize the results in Table~\ref{tab:ablation} to show the effectiveness of the main designs and how we build our method based on the recent SOTA baseline~\cite{valanarasu2022transweather}. It is clear that each design can improve the quality of restored images. Specifically, PSNR loss, SSIM loss and data augmentation are simple and effective ways to improve the performance to be a stronger baseline (defined as $B$). 
The proposed CWP embedding module and SAR encoder with CLIP SRD can further improve the score of PSNR by $0.20$ and $0.28$ respectively, verifying that both weather prior embedding and spatially-adaptive residual architecture trained by knowledge transfer can improve the quality of images restored from adverse weather condition. The visual comparison of these versions is shown in Figure~\ref{fig:vis_bcd}. It is obvious that with the addition of the proposed module, the quality of image restoration is gradually improved, especially as shown in the red rectangle.

\begin{table}[t]
    \centering
    \caption{Comparison of different designs for CWP embedding module. The baseline is the setting of \textit{B} in Table~\ref{tab:ablation}.}
    \begin{tabular}{c|c|c|c|c}
    \toprule
     & Encoder Type & $\mathcal{L}^{c}$ & PSNR $\uparrow$ & SSIM $\uparrow$ \\
    \midrule
    B & $None$ & & 29.23 & 0.9514 \\
    \midrule
    & $learnable$ & & 29.27 & 0.9518 \\
    & $learnable$ & yes & 29.34 & 0.9819 \\
    & $ResNet$ & & 29.28 & 0.9511 \\
    & $ResNet$ & yes & 29.34 & 0.9518 \\
    & $ViT$ & yes & 29.31 & 0.9522 \\
    & $CLIP_{V}$ &  & 29.37 & 0.9521 \\
    \midrule
    & $BLIP_{V}$ & yes & 29.41 & 0.9519 \\
    \midrule
    C & $CLIP_{V}$ & yes & \textbf{29.43} & \textbf{0.9523} \\
    \bottomrule
    \end{tabular}
    \label{tab:ablation_CWP}
    \vspace{-0cm}
\end{table}

\noindent\textbf{CWP Embedding Module.} We study the impact of using different encoders to extract the weather prior for CWP embedding module. In Table~\ref{tab:ablation_CWP}, we compare our CLIP based encoder to two other encoders: 1) the learnable encoder which is built with several convolutions and optimized along with the whole network; 2) the ResNet/ViT encoder which consists of a ImageNet pre-trained ResNet50/ViT-B/32 and two linear layers. The results show that large-scale datasets based CLIP image encoder can improve the PSNR score by $0.14$, while the other encoders achieve marginal improvement (only $0.04$).
With the proposed $\mathcal{L}^{c}$ loss, all encoders can further improve the performance by about $0.06$ in terms of PSNR.
Except CLIP, we also apply another large-scale model, BLIP~\cite{li2022blip} as the encoder. The performance of BLIP is close to CLIP and better than other types of encoders. Among all settings, our CLIP based encoder achieves the best scores. 
The above observations demonstrate the advantage of large-scale vision-language models for weather prior embedding.

\begin{table}[t]
    \centering
    \caption{Comparison of main designs for SAR encoder and CLIP SRD. $T_{d}$ denotes the teacher network, $N_{d}$ means that the distillation loss $\mathcal{L}^{d}$ is introduced from $N_{d}$ epoch. \textit{all} means whether to use all features within each stage of SAR encoder to calculate the loss or only the last one. The baseline is the setting of $C$ in Table~\ref{tab:ablation}.}
    \begin{tabular}{c|c|c|c|c|c|c}
        \toprule
         & \textit{SAR} & $T_{d}$ & $N_{d}$ & \textit{all} & PSNR $\uparrow$ & SSIM $\uparrow$ \\
         \midrule
         C &  &  &  &  & 29.43 & 0.9523 \\
         \midrule
         & yes &  &  &  & 29.64 & 0.9542 \\
         & yes & $ResNet$ & 0 & no & 29.61 & 0.9541 \\
         & yes & $ResNet$ & 0 & yes & 29.64 & 0.9539 \\
         & yes & $CLIP_{R}$ & 0 & no & 29.62 & 0.9536 \\
         & yes & $CLIP_{V}$ & 0 & yes & 29.60 & 0.9539 \\
         \midrule
        D1 & yes & $CLIP_{R}$ & 0 & yes & 29.68 & 0.9540 \\
        D & yes & $CLIP_{R}$ & 200 & yes & \textbf{29.71} & \textbf{0.9544} \\
        \bottomrule
    \end{tabular}
    \label{tab:ablation_sar}
    \vspace{-0.0cm}
\end{table}

\begin{figure}
    \centering
    \includegraphics[width=0.95\linewidth]{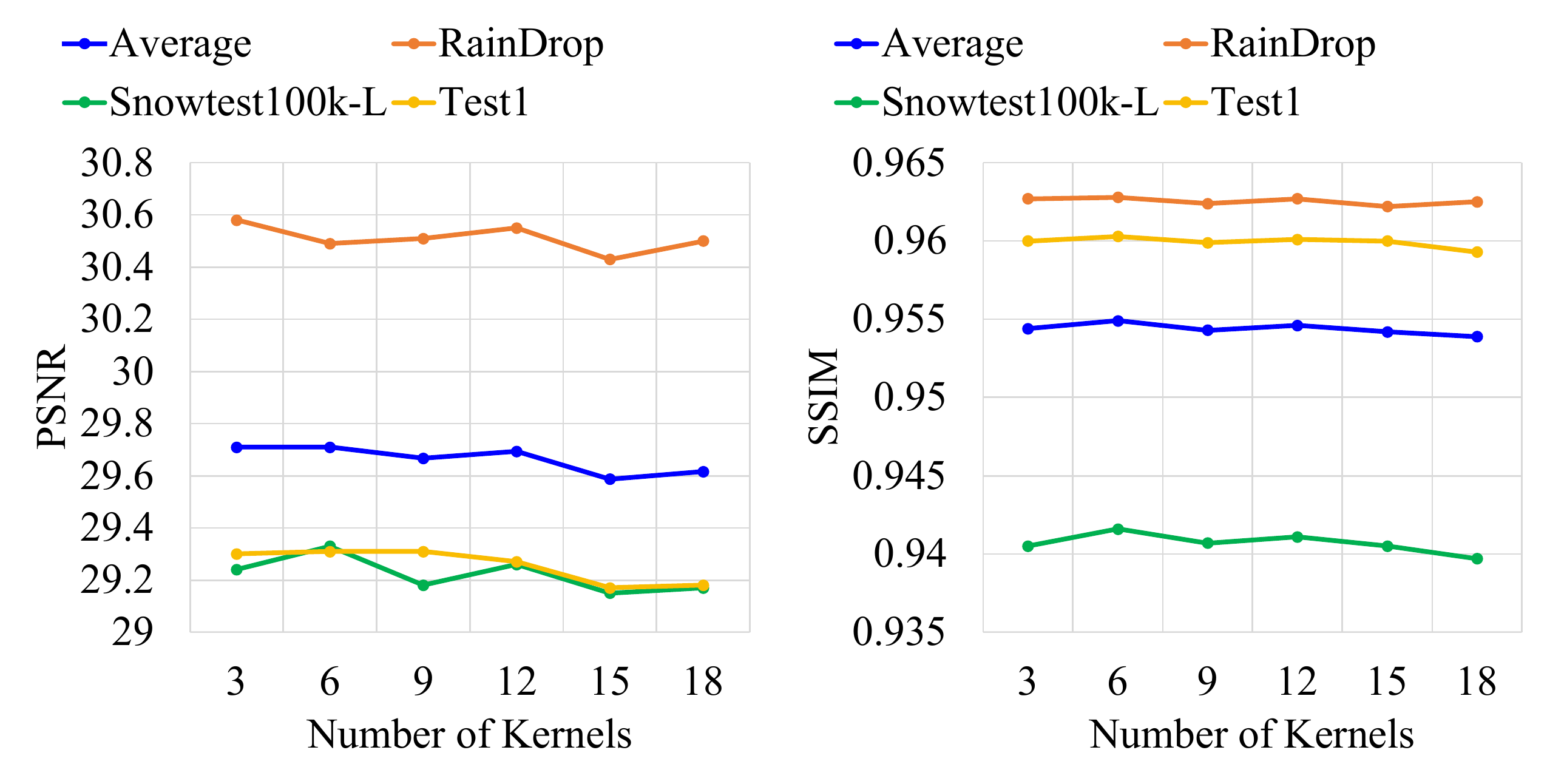}
    \caption{Comparison of different number of kernels in \textit{SAR}. The baseline is the setting of \textit{D} in Table~\ref{tab:ablation}.}
    \label{fig:abl_n}
\end{figure}

\begin{table}[t]
    \centering
    \caption{Comparison of different settings of knowledge transfer. The baseline is the setting of $D1$ in Table~\ref{tab:ablation_sar}.}
    \begin{tabular}{l|c|c}
    \toprule
    Methods & PSNR $\uparrow$ & SSIM $\uparrow$ \\
    \midrule
    D1 & 29.68 & 0.9540 \\
    \midrule
    without residual architecture & 28.67 & 0.9463 \\
    \midrule
    $\mathcal{L}^{d}$ without normalization & 29.32 & 0.9516 \\
    \midrule
    use projection layer & 29.65 & 0.9541 \\
    \bottomrule
    \end{tabular}
    \label{tab:ablation_distill}
    \vspace{-0.0cm}
\end{table}

\begin{table*}
  \centering
\caption{Comparison of MAE as the teacher models for residual knowledge transfer. Other settings are the same to $D1$ in Table~\ref{tab:ablation}.}
  \begin{tabular}{c|c|c|c|c|c|c|c|c}
    \toprule
    \multirow{2}{*}{Teacher models} & \multicolumn{2}{c|}{Snow100K-L~\cite{liu2018desnownet}} & \multicolumn{2}{c|}{Test1~\cite{li2019heavy}} & 
    \multicolumn{2}{c|}{RainDrop~\cite{qian2018attentive}} & \multicolumn{2}{c}{Mean}\\
    \cline{2-9}
     & PSNR$\uparrow$ & SSIM$\uparrow$ & PSNR$\uparrow$ & SSIM$\uparrow$ & PSNR$\uparrow$ & SSIM$\uparrow$ & PSNR$\uparrow$ & SSIM$\uparrow$\\
    \midrule
    $CLIP_{V}$ & 29.04 & 0.9396 & 29.28 & 0.9600 & 30.44 & 0.9619 & 29.59 & 0.9538 \\
    \midrule
    $MAE$ & 29.20 & 0.9396 & 29.24 & 0.9595 & 30.53 & 0.9620 & 29.65 & 0.9537\\
    \bottomrule
  \end{tabular}
  \label{tab:abl_mae}
\end{table*}

\noindent\textbf{SAR Encoder.} In Table~\ref{tab:ablation_sar}, we compare the results with different settings for the proposed SAR encoder. With the proposed SAR architecture, we find that the performance of $C$ is significantly improved (e.g. $0.21$ in terms of PSNR). It means that spatially-adaptive residual mechanism can effectively help to restore areas in different spatial locations. \tzt{In Figure~\ref{fig:abl_n}, we study the number of kernels in \textit{SAR}. Considering the number of weather is three in our experiments, we compare the settings of $N_{dw}=\{3, 6, 9, 12, 15, 18\}$. The results show that $3$ kernels are enough for \textit{SAR} to learn spatially-adaptive residual information. When $N$ is larger than $12$, it seems that the learning becomes difficult, resulting in the degradation of performance.}

\noindent\textbf{CLIP SRD Strategy. }When CLIP is used to transfer knowledge during training, the PSNR score will be further improved to $29.68$ (denoted as $D1$). We also use a ResNet50 pre-trained on ImageNet as the teacher model for comparison, and find that the knowledge seems to have no obvious help. This proves that CLIP can indeed guide the learning. Compared to $CLIP_{V}$ which directly reduces the resolution of features at the beginning, $CLIP_{R}$ gradually reduces that of features during encoding. Its spatial representation of intermediate features is expected to be better, resulting in the improvement on knowledge transfer. In addition, we find that transferring knowledge to all features within each stage is more effective than only to the last one of each stage. It may be caused by the soft residual matching mechanism, which makes the transferred information beneficial to all residual features.

We further discuss some necessary designs for distillation in Table~\ref{tab:ablation_distill}. If we do not use residual architecture in $FFN$ and directly distill the features of CLIP extracted from weather images, the performance will drop sharply ($28.67$ vs. $29.68$ in terms of PSNR). This is because the training objectives of CLIP are different from ours, resulting in a large gap between features. Thus, transferring knowledge directly will seriously hinder the learning of our model. It also explains the importance of normalization in $\mathcal{L}^{d}$. Removing normalization will significantly reduce performance. Besides, it explains why it is more effective to introduce the knowledge of CLIP after several epochs (e.g. 200 epoch) rather than at the beginning of training. Besides, we design a convolution based projection layer to project features into the same space as previous methods, and find that a simple pooling layer is enough to match residual features.

\subsection{\tzt{Discussion about CLIP}}
Although with the well designed CLIP SRD strategy (e.g. residual knowledge transfer and soft feature matching), the improvement brought by CLIP knowledge transfer is still lower than CWP embedding module. It is reasonable since the training objective of CLIP is to align semantics between images and texts. Even though, the promising results in our experiments show that CLIP can still transfer more valuable low-level supervision compared with learnable or ImageNet based models.
This observation inspires us to considering building a pretrained model with not only vision-language semantic representation ability but also low level feature distinctiveness. Such kind of pretrained models may further promote tasks like multi-weather removal that require both semantic and low-level representation. To partially demonstrate it, we use the MAE pretrained model~\cite{he2022masked} as the teacher for local spatial feature representation. MAE trains the encoder via the masked image modeling method, which makes the encoder pay more attention to local spatial representation to reconstruct the image. We use the same ViT-Base network as $CLIP_{V}$ for fair comparison. As summarized in Table~\ref{tab:abl_mae}, MAE performs better than $CLIP_{V}$ in terms of PSNR, even though the pretrained dataset of MAE is much smaller than that of CLIP. It may prove our thinking that a pretrained model focuses on both semantic representation and low level feature distinctiveness can further promote tasks like multi-weather removal.

\section{Conclusions}
In this paper, we focus on the task of removing various adverse weather from images via an unified model. We explore from the aspects of feature representation and semantic embedding under the potential application of large-scale vision-language models (CLIP). 
First, we propose SAR encoder to enhance features through spatially-adaptive residual architecture and guide the training of these residual features via the knowledge from CLIP between clean and weather images with the proposed CLIP SRD strategy. Then, we embed CLIP weather prior together with a learnable weather prior through the proposed CWP embedding module. Experiments demonstrate the effectiveness of our method.

\bibliographystyle{IEEEtran}
\bibliography{egbib}

\end{document}